\documentclass[letterpaper]{article} 
\usepackage[draft]{aaai2026}

\usepackage{times}  
\usepackage{helvet}  
\usepackage{courier}  
\usepackage[hyphens]{url}  
\usepackage{graphicx} 
\usepackage{natbib}  
\usepackage{caption} 
\usepackage{algorithm}
\usepackage{algorithmic}
\usepackage{microtype}
\usepackage{amsmath}
\usepackage{arydshln}
\usepackage{booktabs}
\usepackage{multirow}
\usepackage{amssymb}

\usepackage{pifont}

\usepackage{newfloat}
\usepackage{listings}
\DeclareCaptionStyle{ruled}{labelfont=normalfont,labelsep=colon,strut=off} 
\floatstyle{ruled}
\newfloat{listing}{tb}{lst}{}
\floatname{listing}{Listing}
\title{Self-Supervised Cross-Modal Learning for Image-to-Point Cloud Registration }

\author{
    Xingmei Wang\textsuperscript{\rm 1},
    Xiaoyu Hu\textsuperscript{\rm 1},
    Chengkai Huang\textsuperscript{\rm 2,3}\thanks{Corresponding author},
    Ziyan Zeng\textsuperscript{\rm 1}, \\
    Guohao Nie\textsuperscript{\rm 1},
    Quan Z. Sheng\textsuperscript{\rm 2},
    Lina Yao\textsuperscript{\rm 3,4}
}
\affiliations{
    \textsuperscript{\rm 1}Harbin Engineering University,
    \textsuperscript{\rm 2}Macquarie University,
    \textsuperscript{\rm 3}University of New South Wales,
    \textsuperscript{\rm 4}CSIRO's Data61 \quad \\
    \{wangxingmei, xiaoyuhu, z13204639269, nieguohao\}@hrbeu.edu.cn,\\
    michael.sheng@mq.edu.au,
    chengkai.huang@mq.edu.au,
    lina.yao@unsw.edu.au
}

\usepackage{bibentry}

\begin{document}

\maketitle

\begin{abstract}

Bridging 2D and 3D sensor modalities is critical for robust perception in autonomous systems. However, image-to-point cloud (I2P) registration remains challenging due to the semantic–geometric gap between texture-rich but depth-ambiguous images and sparse yet metrically precise point clouds, as well as the tendency of existing methods to converge to local optima. To overcome these limitations, we introduce CrossI2P, a self-supervised framework that unifies cross-modal learning and two-stage registration in a single end-to-end pipeline.
First, we learn a geometric-semantic fused embedding space via dual-path contrastive learning, enabling annotation-free, bidirectional alignment of 2D textures and 3D structures. Second, we adopt a coarse-to-fine registration paradigm: a global stage establishes superpoint–superpixel correspondences through joint intra-modal context and cross-modal interaction modeling, followed by a geometry-constrained point-level refinement for precise registration. Third, we employ a dynamic training mechanism with gradient normalization to balance losses for feature alignment, correspondence refinement, and pose estimation.
Extensive experiments demonstrate that CrossI2P outperforms state-of-the-art methods by 23.7\% on the KITTI Odometry benchmark and by 37.9\% on nuScenes, significantly improving both accuracy and robustness.

\end{abstract}




\maketitle

\section{Introduction}

Accurate 2D–3D registration is essential for aligning data from modalities such as cameras and LiDAR to a unified representation, ensuring geometric consistency across sources and enabling effective multimodal fusion for robust scene understanding and decision-making in autonomous driving and robotic perception systems. However, practical deployment on mobile platforms, handheld mapping devices, UAV-mounted sensors, or retrofitted vehicles, often lacks pre-calibrated extrinsics due to mechanical vibrations, thermal variations, ad-hoc sensor reconfiguration, or calibration drift in long-term or unstructured operation, challenges that persist even when benchmarks like Waymo \cite{1} and KITTI \cite{2} assume known sensor setups. Moreover, the inherent cross-modal representation gap, images offer dense textures but lack depth, while point clouds provide sparse geometry with limited semantics, leads to divergent feature emphases, with images focusing on semantic attributes \cite{3,4,5} and point clouds prioritizing geometric structures \cite{6,7}, highlighting the need for calibration-free, robust 2D–3D registration methods. Nonetheless, current 2D–3D registration approaches face some significant challenges.

\begin{figure}[t]
	\centering 
	\includegraphics[width=1.0\hsize]{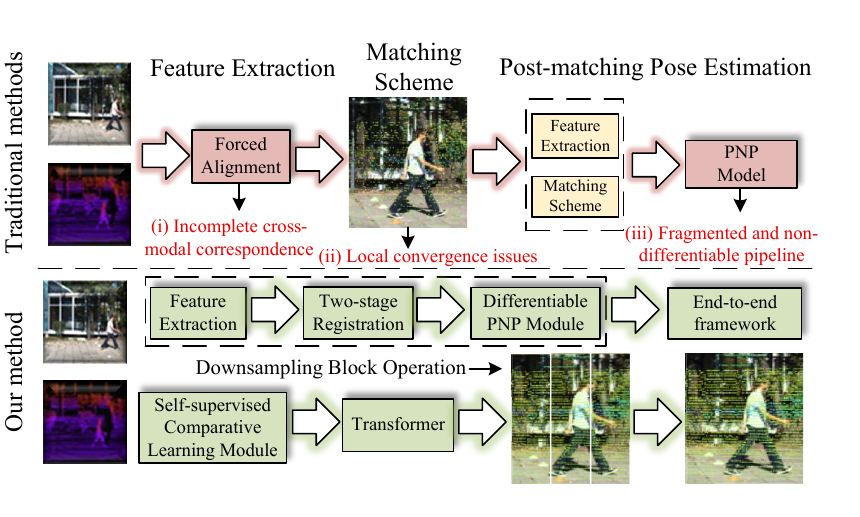}
    \caption{Comparison of traditional image-to-point cloud registration pipelines (top) versus our CrossI2P framework (bottom). 
    }
	\label{fig:introduction}
\end{figure}


First, the inherent disparity between 2D image and 3D point cloud feature representations creates a large semantic gap that hinders direct matching. Methods like DeepI2P \cite{40} fuse multi-level features but can be computationally intensive and still struggle in textureless or ambiguous regions, while CoFiI2P \cite{42} injects high-level semantics to filter outliers at the cost of overlooking fine geometric details. Self-supervised contrastive frameworks such as CLIP \cite{14}, achieve strong semantic alignment but remain unaware of the geometric constraints critical for registration. Despite progress, \textit{\textbf{(Gap 1)} a semantic mismatch persists between cross-modal pre-trained features and those required for accurate geometric registration, making it difficult to fully leverage large-scale cross-modal pre-training for precise registration tasks.}



Second, traditional registration pipelines depend on local feature correspondences (such as SIFT, ORB, or learned descriptors) followed by robust pose estimation (e.g., RANSAC + Perspective-n-Point (PnP)). Because these methods ignore the global scene structure, they often fall into local minima or yield incorrect poses when faced with strong outliers or visually ambiguous regions. To inject broader context, some approaches retrieve database images to supply additional viewpoints \cite{34}, use visibility priors to favor geometrically plausible matches \cite{37,38}, apply 2D–3D vocabulary-based retrieval for global feature indexing \cite{39}, or incorporate CoFiI2P’s coarse-to-fine semantic filtering for outlier rejection \cite{42}. 
However, these context-aware strategies often rely on large external image repositories, can suffer from inaccurate visibility estimation under occlusions or sensor noise, use fixed vocabularies that limit coverage of diverse feature variations, and even coarse-to-fine methods like CoFiI2P still struggle in low-density regions. \textit{\textbf{(Gap 2)} Effectively modeling and leveraging the full relational topology of large, complex scenes remains an open challenge for robust, calibration-free registration.}


Third, enabling end-to-end learning is blocked by the Perspective-n-Point (PnP) solver, a core component of traditional pipelines that is inherently non-trainable. As a result, gradients from pose estimation cannot flow back through PnP to feature extraction and matching stages, preventing joint optimization for pose accuracy. Most deep learning methods, such as DeepI2P \cite{40} and CoFiI2P \cite{42}, sidestep this by applying classification or contrastive losses directly to features or match scores, then treating PnP as a fixed post-processing step. Consequently, \textit{\textbf{(Gap 3)} no practical, differentiable substitute for PnP exists that would allow true end-to-end co-optimization within large-scale 2D–3D registration pipelines.}

To bridge these gaps, we propose CrossI2P, a self-supervised framework for image-to-point cloud registration. As shown in Figure \ref{fig:introduction}, first, to close the semantic divide between 2D images and 3D point clouds, we design a self-supervised comparative learning module that projects 3D points into the camera frame and pulls positive image–point pairs together in feature space via a contrastive loss. Second, to escape local minima and capture global context, we introduce a two-stage registration pipeline: a coarse matching stage aligns superpoints with superpixels using region-level descriptors, then a fine stage refines these correspondences by aligning local patches within each region. Finally, to enable end-to-end optimization, we integrate a differentiable PnP layer into the network and harmonize losses across modules with a dynamic collaborative training scheme. CrossI2P thus achieves semantic-geometric alignment, global consistency, and full differentiability for accurate and robust 2D–3D registration.
The main contributions of this work are as follows:
\begin{itemize}


\item We introduce CrossI2P, a unified self-supervised framework designed for accurate and robust 2D–3D registration.

\item Within this framework, we develop three core components: a self-supervised comparative learning module for feature alignment, a coarse‐to‐fine registration strategy based on superpoint–superpixel matching, and a differentiable PnP solver coordinated via dynamic gradient collaboration.

\item Extensive experiments on KITTI and nuScenes demonstrate that CrossI2P consistently achieves state‐of‐the‐art performance under diverse data conditions.



\end{itemize}

\section{Related Work}
\label{sec:related}
\subsection{Image-Point Cloud Registration Scheme}
Traditional 2D–3D registration evolved from small‐scale SLAM systems \cite{30,31,32,33} to large‐scale image retrieval frameworks that match 2D features against reconstructed and synthetic views \cite{34}, accelerated by GPU‐based vocabulary trees \cite{35} and adapted for MAV localization \cite{36}. Visibility‐aware strategies then learned to predict or prioritize visible 3D points \cite{37,38}, and Sattler et al. introduced direct 2D–3D vocabulary matching for greater efficiency \cite{39}. Deep learning fused modalities: DeepI2P \cite{40} classifies points in camera frustums, CorrI2P \cite{41} performs dense overlap matching but lacks global context, and CoFiI2P \cite{42} injects semantic cues in a coarse‐to‐fine pipeline with global optimization. However, CoFiI2P’s reliance on traditional feature extraction and comparison still leaves the semantic gap between 2D and 3D features insufficiently addressed.
\subsection{Self-supervised Cross-Modal Learning}
Contrastive Multi-view Coding (CMC) \cite{10}, AVSA \cite{11}, and MMV \cite{12} pioneered self-supervised alignment across different views and modalities. This approach scaled massively with CLIP \cite{14} and ALIGN \cite{15}, which leverage over 400 M image–text pairs, and gained efficiency via CLIPPO’s pixel unification \cite{16} and FLIP’s masked patch training \cite{17}. Intra-modal contrastive losses, as in SLIP \cite{18}, CrossCLR \cite{19}, and CrossPoint \cite{20}, further reinforce alignment—though AVID \cite{21} warns of low-level statistic dominance. Cross-modal matching spans audio-visual correspondence \cite{22,23}, temporal-hard negatives \cite{24}, and pixel-level separation \cite{25,27}, while image–text matching methods like UNITER \cite{28} and ALBEF \cite{29} introduce hard negatives to improve discrimination. Unlike these prior works, our approach reconciles semantic alignment with geometric precision and avoids local optima through a dedicated two-stage registration pipeline.

\begin{figure*}[!htb]
	\centering 
	\includegraphics[width=1.0\hsize]{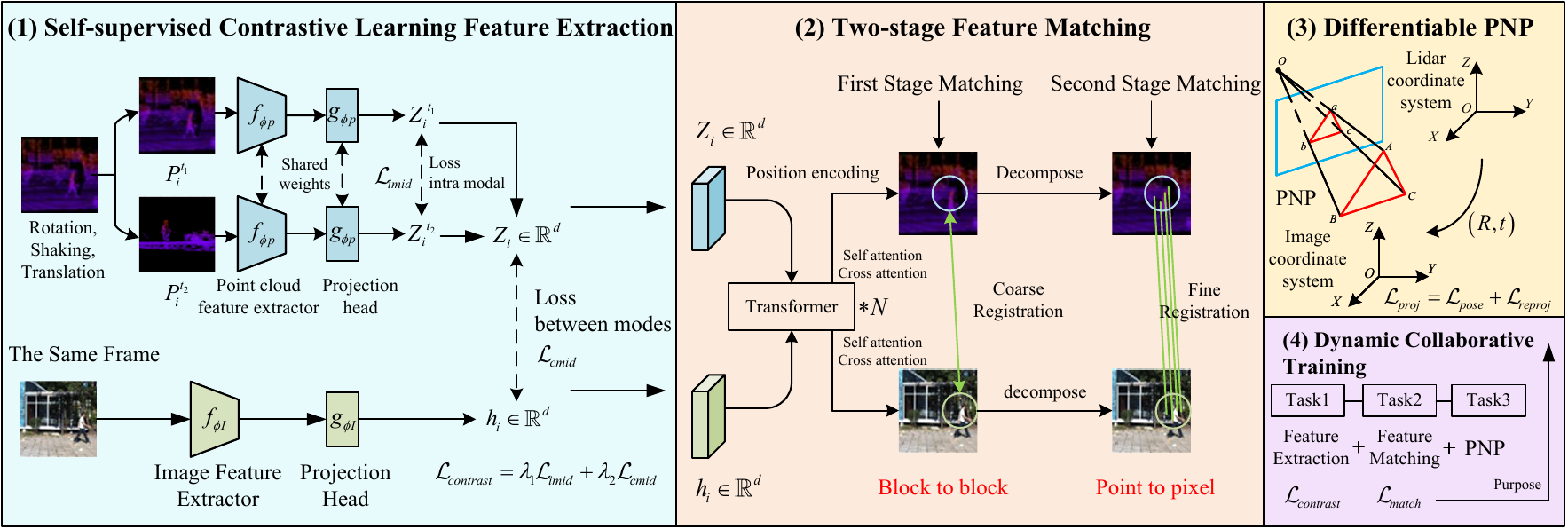}
	\caption{CrossI2P: Firstly, a Self-supervised contrastive Learning network (SL) is used to map the two modalities into a shared feature space. Subsequently, two-stage Registration feature Matching (RM) is implemented to obtain the final correspondence relationship. The entire process is optimized end-to-end through backpropagation using the Differentiable PNP algorithm (D-PNP). Finally, it was optimized using the Dynamic Collaborative training (DC) method.}
	\label{fig:overview}
\end{figure*}


\section{Method}\label{sec:method}


Given an overlapping image-point cloud pair, denoted as $I \in \mathbb{R}^{3 \times W \times H}$ and $P \in \mathbb{R}^{N \times 3}$, where $W$ and $H$ represent the image width and height, and $N$ denotes the number of 3D points, the goal of image-to-point cloud registration is to establish accurate correspondences between image pixels and 3D points in the point cloud. This enables the estimation of the relative transformation between the two modalities, including both rotation and translation components. 

To this end, we propose CrossI2P, a framework consisting of four key components: (1) a self-supervised contrastive learning module for cross-modal feature alignment, (2) a feature registration module for hierarchical matching, (3) an attitude estimation module for camera pose regression, and (4) a Dynamic Collaborative Training mechanism to enhance robustness.

\subsection{Architecture Overview}

As shown in Figure \ref{fig:overview}, we propose a two-stage image-point cloud registration framework based on self-supervised learning for cross-modal registration under complex perception scenarios. Specifically, given multi-sensor image and point cloud data from the same scene, we first employ a Self-supervised contrastive Learning network (SL) to map both modalities into a shared feature space, facilitating cross-modal feature alignment and extracting common representations. Subsequently, a two-stage Registration feature Matching (RM) is implemented: The initial stage utilizes a transformer network for global feature matching to obtain region-based coarse registration results, followed by point-to-point registration of internal elements within these regions to derive final correspondence. The entire process is end-to-end optimized through backpropagation using a Differentiable PNP algorithm (D-PNP), which addresses the non-differentiable limitation of traditional Perspective-n-Point methods while refining camera poses. Finally, it is optimized by the Dynamic Collaborative training method (DC). 

\subsection{Self-supervised Comparative Learning Module} \label{subsection:Self supervised comparative learning module}
\begin{figure}[t]
	\centering 
	\includegraphics[width=1.0\hsize]{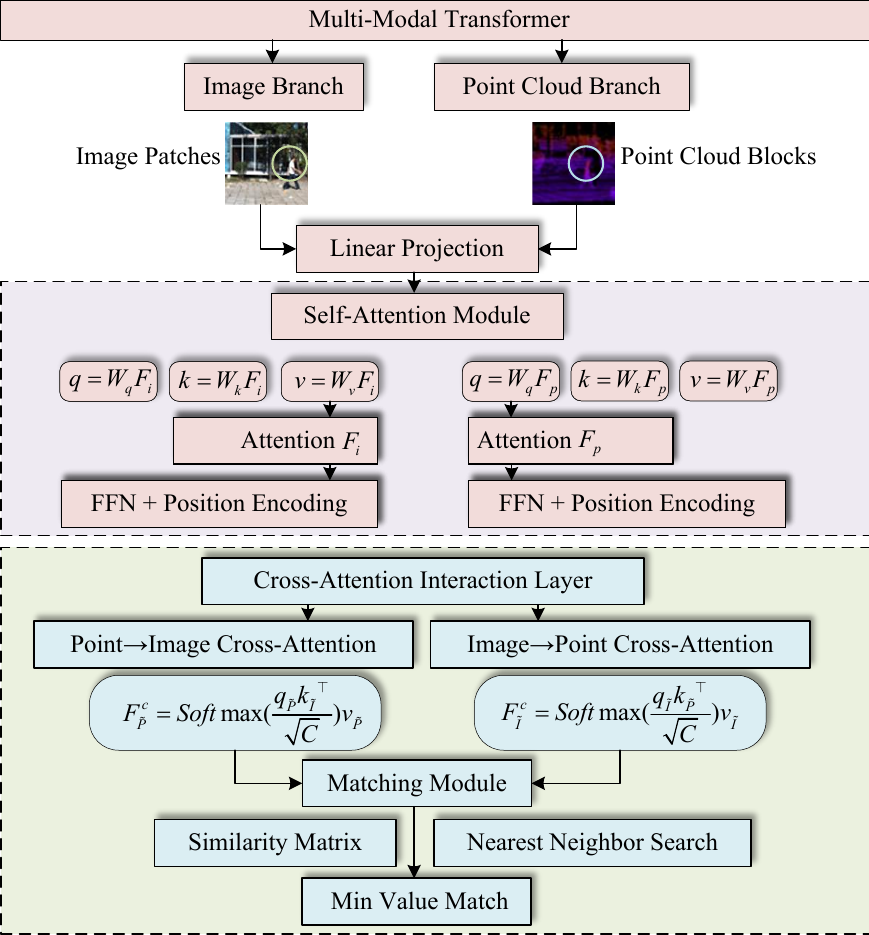}
	\caption{ Overview of the two-stage feature matching module in CrossI2P.  }
    
	\label{fig:Feature registration module}
\end{figure}

Firstly, we develop the intra-modal instance discrimination (IMID) to implement invariance for a set of point cloud geometric transformations $T$ by performing self-supervised comparative learning. 

As shown in Figure \ref{fig:Feature registration module}, the point cloud feature extractor ${f_{\phi p}}$  projects two point clouds into the feature embedding space, and uses a specific projection head ${g_{\phi p}}$  to project the obtained feature vector into the invariant space ${{\mathbb R}^d}$with contrast loss applied, we express the projection vectors of ${P_i}^{{t_1}}$ and ${P_i}^{{t_2}}$ as ${Z_i}^{{t_1}}$ and ${Z_i}^{{t_1}}$, respectively where 
${Z_i}^t = {g_{\phi p}}({f_{\phi p}}({P_i}^t))$.The goal here is to maximize the similarity between ${Z_i}^{{t_1}}$ and ${Z_i}^{{t_1}}$ while minimizing the similarity with all other projection vectors in a small batch of point clouds. We calculate the loss function 
$l(i,{t_1},{t_2})$ for the examples ${Z^{{t_1}}}$ and ${Z^{{t_2}}}$ as:

\begin{equation}
l(i,t_1,t_2)
= -\log
  \frac{
    \exp\bigl(s(z_i^{t_1},z_i^{t_2})/\tau\bigr)
  }{
    \displaystyle
    \begin{array}{@{}l@{}}
      \sum_{k\neq i}\exp\bigl(s(z_i^{t_1},z_k^{t_1})/\tau\bigr)\\[0.5ex]
      +\;\sum_{k=1}^N\exp\bigl(s(z_i^{t_1},z_k^{t_2})/\tau\bigr)
    \end{array}
  },
\end{equation}
where $N$ is the minimum batch size, $\tau $ is the temperature coefficient, and $s\left( {{\bf{z}}_i^{{t_1}},{\bf{z}}_k^{{t_1}}} \right)$ is the cosine similarity function. Our modal case discrimination loss function ${\mathcal{L}}_{\text{imid}}$ can be described as:
\begin{equation}
{\mathcal L}_{\text{imid}} = \frac{1}{2N} \sum\limits_{i=1}^N \left[ l\left( i, t_1, t_2 \right) + l\left( i, t_2, t_1 \right) \right],
\end{equation}
In addition to the feature alignment within the point cloud mode, we also introduce auxiliary contrast targets between the point cloud and the image mode to learn the distinguishing features, so as to obtain better learning ability of 3D point cloud representation.
To this end, we first use the visual backbone ${f_{\phi I}}$ to embed the 2D image I into the feature space. The commonly used RESNET architecture is chosen as ${f_{\phi I}}$. Then, the feature vector is projected into the invariant space ${{\mathbb R}^d}$ using the image projection head ${g_{\phi I}}$. Projected image features are defined as 
${h_i}$ ,where: 
\begin{equation}
{h_i} = {g_{\phi I}}({f_{\phi I}}({I_i})),
\end{equation}
In order to reduce the influence of point cloud rigid body transformation, we calculate the average value of the projection vectors 
${Z_i}^{{t_1}}$ and ${Z_i}^{{t_2}}$ to obtain the projection prototype vector ${Z_i}$ of ${P_i}$:
\begin{equation}
{Z_i} = \frac{1}{2}({Z_i}^{{t_1}} + {Z_i}^{{t_2}}),
\end{equation}
We aim to maximize the similarity of ${Z_i}$ and ${h_i}$ in invariant space because they all correspond to the same object. Our cross-modal alignment forced model learns from more difficult positive and negative samples, thus enhancing the representation ability than only learning from intra-modal alignment. The loss function 
$l(i,z,h)$ of positive samples for ${Z_i}$ and ${h_i}$ is:
\begin{equation}
c(i, \mathbf{z}, \mathbf{h}) = -\log \frac{
    \exp\left( s(\mathbf{z}_i, \mathbf{h}_i) / \tau \right)
}{
    \begin{array}{@{}c@{}}
      \sum\limits_{\substack{k=1 \\ k \neq i}}^N 
      \exp\left( s(\mathbf{z}_i, \mathbf{z}_k) / \tau \right) 
      \\[-2.5ex]  
      \quad + \quad
      \sum\limits_{k=1}^N \exp\left( s(\mathbf{z}_i, \mathbf{h}_k) / \tau \right)
    \end{array}
},
\end{equation}
where $N$ is the minimum batch size,
$\tau$ is the temperature coefficient, and $s\left( {{\bf{z}}_i^{{t_1}},{\bf{z}}_k^{{t_1}}} \right)$ is the cosine similarity function. The trans-modal loss function ${\mathcal{L}}_{\text{cmid}}$ is formulated as:
\begin{equation}
{{\mathcal L}_{cmid}} = \frac{1}{{2N}}\mathop \sum \limits_{i = 1}^N \left[ {c\left( {i,z,h} \right) + c\left( {i,h,z} \right)} \right],
\end{equation}
The final loss function is defined as
${{\mathcal L}_{contrast}}$:
\begin{equation}
{{\mathcal L}_{contrast}} = {\lambda _1}{{\mathcal L}_{imid}} + {\lambda _2}{{\mathcal L}_{cmid}},
\end{equation}
where ${\mathcal{L}}_{\text{imid}}$ imposes invariance on the point cloud transformation, while ${\mathcal{L}}_{\text{cmid}}$ injects 3D-2D correspondence.
\subsection{Two-stage Registration Feature Matching} 
\label{subsection:Feature registration module}

In the feature registration module, the transformer is used to obtain the geometric and spatial consistency between the image and the point cloud. Each level of the transformer includes a self-attention module for long-distance context between channels and a cross-attention module for feature exchange within channels. 
First, we construct local point blocks and image pixel blocks according to the distribution of points in space and the feature pyramid. Local point blocks are defined as $\tilde P$, and image pixel blocks are defined as $\tilde I$.T he independent points and pixels are defined as  ${\tilde P_i}$ and ${\tilde I_i}$.

\begin{equation}
\tilde P = \left\{ {p \in P\left| {\left\| {p - {{\tilde p}_i}} \right\| < r} \right.} \right\},
\end{equation}
where $r$ is the selected radius. Image pixel blocks use a pyramid matching strategy to match the multi-scale features obtained by the image pyramid within the layer and cross-layer weighted fusion to obtain the final local pixel block $\tilde I$.
\subsubsection{First stage registration.} Our transformer contains both a self-attention module for spatial context capture in homogeneous data and a cross-attention module for mixed feature extraction in heterogeneous data.
For the self-attention module, given the coarse feature map $F \in {{\mathbb R}^{N \times C}}$ of the image or point cloud, the query vector, key vector and value vector $Q$, $K$, $V$ are generated as:
\begin{equation}
q = {W_q}F,k = {W_k}F,v = {W_v}F,
\end{equation}
where ${W_q}$,${W_k}$,${W_v}$ is a learnable weight matrix. Then, the global attention enhancement feature map is calculated as:
\begin{equation}
{F^s} = \text{Softmax}(\frac{{q{k^ \top }}}{{\sqrt C }})v,
\end{equation}

The global perceptual feature $f$ extracted from the channel dimension is fed into the feedforward network (FFN) to fuse the spatial relationship information. Given the feature graph ${F^s}$, the relative position is encoded by an MLP.
Cross attention is designed to fuse image and point cloud features.
Given the feature map 
${F_{\tilde{P}}}$ of the local point blocks $\tilde{P}$ and ${F_{\tilde I}}$ of the local point blocks $\tilde I$.The cross attention enhancement feature map ${F_{\tilde P}}^c$ of the point cloud and ${F_{\tilde I}}^c$of the image are represented as:
\begin{equation}
F_{\tilde P}^c = \text{Softmax}(\frac{{{q_{\tilde P}}{k_{\tilde I}}^ \top }}{{\sqrt C }}){v_{\tilde P}},
\end{equation}
\begin{equation}
F_{\tilde I}^c = \text{Softmax}(\frac{{{q_{\tilde I}}{k_{\tilde P}}^ \top }}{{\sqrt C }}){v_{\tilde I}},
\end{equation}
where ${q_{\tilde P}}$, ${k_{\tilde P}}$ and ${v_{\tilde P}}$ are query vectors, key vectors and value vectors of point cloud feature ${F_{\tilde{P}}}$, ${q_{\tilde I}}$, ${k_{\tilde I}}$ and ${v_{\tilde I}}$
are query vectors, key vectors and value vectors of point cloud feature ${F_{\tilde{I}}}$.
Then, the coarse-level correspondence between the local point blocks and the image pixel blocks is estimated by sorting the nearest image pixel blocks. Matching succeeded when $
{{\mathcal L}_{match}}=\left\| {{F_{\tilde P}} - {F_{\tilde I}}} \right\|$ taking the minimum value.
\subsubsection{Second stage registration.}
Based on the registration results of the first stage, the second stage carries out point-to-point registration in the registered local point blocks and image pixel blocks.
local point blocks $\tilde P$ and image pixel blocks $\tilde I$ inverse decoded into points ${\tilde P_i}$
and  pixels ${\tilde I_j}$, Matching succeeded when $
{{\mathcal L}_{match}}=\left\| {{F_{\tilde P}} - {F_{\tilde I}}} \right\|$ taking the minimum value.

\subsection{Differentiable PNP Module}
\label{subsection:D-PNP}

The differential PNP algorithm embeds the PNP solving process into the deep learning framework, and the gradient can be backpropagated to the front-end feature extraction network or 3D point cloud generation network to achieve joint optimization. For previously obtained 3D point and pixel pairs $\left\{ {\left( {{P_i},{I_i}} \right)} \right\}$ ,where ${P_i}$ is 3D point, ${I_i}$ is its corresponding 2D projection element.Using the closed form method such as EPNP, the gradient of pose parameters $\left( {R,t} \right)$ is calculated by re-parameterization or implicit derivation. The solution formula is:
\begin{equation}
\arg\min {\sum\limits_{\rm{i}} {\left\| {\pi \left( {R{P_{\rm{i}}} + {\rm{t}}} \right) - {I_{\rm{i}}}} \right\|} ^2},
\end{equation}
The loss calculation includes two parts: pose loss and re-projection error loss. Pose loss direct comparison between the predicted pose and the true value.
\begin{equation}
{{\mathcal L}_{pose}} = {\left\| {{R_{pred}} - {R_{gt}}} \right\|^2} + \lambda {\left\| {{t_{pred}} - {t_{gt}}} \right\|^2},
\end{equation}
Indirect supervision of re-projection error loss.
\begin{equation}
{{\mathcal L}_{{\rm{reproj}}}} = {\sum\limits_{\rm{i}} {\left\| {\pi \left( {{R_{{\rm{pred}}}}{P_{\rm{i}}} + {{\rm{t}}_{{\rm{pred}}}}} \right) - {I_{\rm{i}}}} \right\|} ^2},
\end{equation}
\begin{equation}
{{\mathcal L}_{proj}} = {{\mathcal L}_{pose}} + {{\mathcal L}_{reproj}},
\end{equation}
Then the gradient of pose loss or re-projection error is transmitted back to the feature extraction network through the differentiable PNP module.

\subsection{Dynamic Collaborative Training}
\label{subsection:Dynamic Collaborative Training}

The role of the Dynamic Collaborative Training process is to improve the generalization ability, robustness and data efficiency of the model by jointly optimizing multiple related tasks, making full use of the synergy effect between tasks. Its core value is to achieve complementarity and balance among multiple tasks through dynamic adjustment of task weights, shared feature representation and end-to-end optimization.
Image to point cloud registration consists of two tasks: self-supervised feature extraction and feature matching and differentiable PNP. The parameters of the first two tasks are adjusted with the third task as the overall goal.
\begin{equation}
{{\mathcal L}_{total}} = {\lambda _1}{{\mathcal L}_{contrast}} + {\lambda _2}{{\mathcal L}_{match}} + {\lambda _3}{{\mathcal L}_{proj}},
\end{equation}
where $\lambda$ is the dynamic weight, adjusted by GradNorm \cite{44}. differentiable PNP module: Adjust parameters in the solution process (such as iteration step size). Feature matching module: Optimize matching configuration reliability calculation. Self supervised feature extraction module: improves the adaptability of features to pose estimation.

\section{Experiment Settings}
\label{sec:expset}

\subsection{Dataset Preparation}
We conducted comprehensive evaluations on two authoritative autonomous driving benchmarks, Kitti Odometry \cite{48} and nuScenes \cite{49}, where Kitti provides precisely calibrated multi-sensor data with centimeter-accurate 6-DoF ground truth across 39.2 km of urban driving sequences, while nuScenes extends validation to 1,000 multimodal scenarios featuring 6 cameras, 32-beam LiDAR, 5 radars, collectively demonstrating our method's superior robustness across heterogeneous driving conditions and sensor configurations.
\subsection{Evaluation Metrics}

Following previous works \cite{41,42}, we calculate the relative rotation error (RRE), relative translation error (RTE), and registration recall (RR) to evaluate the registration results. RRE and RTE are defined as:
\begin{equation}
\mathrm{RRE} = \sum\limits_{i = 1}^3 {\left| {r\left( i \right)} \right|},
\end{equation}
\begin{equation}
\mathrm{RTE} = \left\| {{t_{gt}} - {t_{pre}}} \right\|,
\end{equation}
where $r$ is the Euler angle vector of $R_{gt}^{ - 1}{R_{pre}}$,  ${R_{gt}}$ and ${t_{gt}}$ are the true rotation and translation matrices of the ground, ${R_{pre}}$ and ${t_{pre}}$are the estimated rotation and translation matrices.
The IR index matches the internal registration accuracy, which measures the specific registration quality of each frame. If the corresponding point cloud pixel pairs in a certain frame are X pairs, then IR is defined as:
\begin{equation}
\mathrm{IR} = \frac{1}{X}\sum\limits_{\left( {{P_i},{I_j}} \right)} {1\left( {\left\| {{\Gamma _{{P_i}}} - {I_j}} \right\| < \tau } \right)},
\end{equation}
where $1\left(  \cdot  \right)$ indicates the function, output 1 if the condition is met, otherwise 0. ${\Gamma _{{P_i}}}$ represents the true coordinates projected onto the image through the projection matrix. $\tau $ is used to control the re-projection error tolerance.
We define RMSE to represent the matching quality of the entire point cloud image pair:
\begin{equation}
\mathrm{RMSE} = \sqrt {\frac{1}{X}\sum\limits_{\left( {{P_i},{I_j}} \right)} {\left\| {{\Gamma _{{P_i}}} - {I_j}} \right\|} },  
\end{equation}
The percentage of correctly matched pairs estimated by the RR metric indicates the descriptor learning ability of the network, which can measure how many correct pairs the algorithm finds in image and point cloud matching tasks. In other words, this means looking at how many reliable matches the algorithm has found. Assuming the number of point cloud image data pairs for registration is Z. RR is defined as:
\begin{equation}
\mathrm{RR} = \frac{1}{Z}\sum\limits_{{\rm{i}} = 1}^Z {1\left( {RMSE < \tau } \right)}. 
\end{equation}
\subsection{Implementation Details}
In the process of detailed experiments, we cut the image to 160 $\times$512 size, down-sampled the point cloud with 0.1M $\times$0.1M $\times$0.1M voxels, and then randomly sampled 20480 points as input. In the self-monitoring module, we use DGCNN \cite{46} as the point cloud feature extractor and ResNet50 \cite{47} as the image feature extractor and debug the baseline to achieve the best effect. The two-layer MLP is used as the projection head to generate 256-dimensional feature vectors projected in the invariant space ${{\mathbb R}^d}$.
The point cloud feature extractor ${f_{{\theta _P}}}\left(  \cdot  \right)$ and image feature extractor ${f_{{\theta _I}}}\left(  \cdot  \right)$ are obtained through training. During the training process, the correct transformation parameters provided by the calibration file are used to establish the ground-truth value correspondence to monitor the network. We trained the entire network with 100 epochs and the batch size was 1. We use Adam to optimize the network. The initial learning rate is 0.001, multiplied by 0.25 after every 5 epochs. For our combined loss, we set $\lambda  = 0.5$. The confidence threshold for successful matching is set to 0.9; that is, when 90\% of the points are correctly matched, we believe that the matching of this frame is reliable.
The whole training was conducted on two Nvidia-Titan 24GB GPUs. For the reproducibility of the results, our detailed code and implementation process will be uploaded to the following website for learning and communication.
 \subsection{Experimental Result }
In order to ensure the fairness of the comparison of experimental results, we used the Kitti odometry data set for the experiments. The main algorithms for comparison include traditional methods(SIFT+PNP) , P2net+PNP \cite{45}, Deepi2p \cite{40}, CorrI2P \cite{41}, CoFiI2P\cite{42}. In order to compare with the most advanced registration methods available, we set various threshold parameters as the same as possible. For example, we set a set of specific thresholds  $\left( {10^\circ /5m} \right)$ to reject the wrong registration frame. The experimental results are shown in Table 1.

\begin{table*}[ht]
\small
\centering
\fontsize{9pt}{10pt}\selectfont
    \setlength{\tabcolsep}{2.5pt}
\caption{Experimental results on the KITTI and nuScenes datasets, with the best-performing values highlighted in bold.}
\label{tab:combined_results}
\begin{tabular}{lcccccccc}
\toprule
\multirow{2}{*}{\textbf{Method}}
  & \multicolumn{4}{c}{\textbf{KITTI}}
  & \multicolumn{4}{c}{\textbf{nuScenes}} \\
\cmidrule(lr){2-5}\cmidrule(lr){6-9}
 & \textbf{RRE$\downarrow$($^\circ$)} & \textbf{RTE$\downarrow$(m)} & \textbf{IR(\%)} & \textbf{RR(\%)}
 & \textbf{RRE$\downarrow$($^\circ$)} & \textbf{RTE$\downarrow$(m)} & \textbf{IR(\%)} & \textbf{RR(\%)} \\
\midrule
SIFT+PNP
  & 12.32±18.13 & 3.13±2.96 & –    & –    
  & 15.62±10.12 & 5.13±4.37 & –    & – \\
P2net+PNP
  & 10.35±8.47  & 3.11±3.55 & –    & –    
  & 11.25±7.27  & 4.16±3.64 & –    & – \\
Deepi2p
  & 15.52±12.73 & 3.17±3.22 & –    & –    
  & 13.52±12.73 & 3.97±2.35 & –    & – \\
Corri2p
  & 2.70±1.97   & 1.24±0.87 & 25.84 & 90.66
  & 8.70±1.97   & 3.24±1.86 & 26.87 & 88.57 \\
CoFiI2P
  & 1.14±0.78   & 0.29±0.19 & 75.75 & 100.00
  & 5.14±0.78   & 2.29±1.19 & 73.60 & 100.00 \\
Our method
  & \textbf{0.87±0.33 (–23.6\%)} & \textbf{0.18±0.11 (–37.9\%)}
      & \textbf{88.74} & \textbf{100.00}
  & \textbf{3.92±2.29 (–23.7\%)} & \textbf{1.74±0.98 (–24.0\%)}
      & \textbf{82.58} & \textbf{100.00} \\
\bottomrule
\end{tabular}
\end{table*}

From Table \ref{tab:combined_results}, it can be seen that when the threshold is ($10^\circ/5m$), our method improves IR while reducing RRE and RTE errors and minimizing fluctuations, indicating that the self-supervised comparative learning feature extraction method plays a very positive role in cross-modal feature matching.
To better reflect the advantages of our algorithm, we use histograms to display the stability of RRE and RTE parameters for different methods, as shown in Figure \ref{fig:View}. The height of the column indicates the size of the parameter error, and the lower the value, the smaller the overall error.
\begin{figure}[!htb]
	\flushleft 
	\includegraphics[width=1.0\hsize]{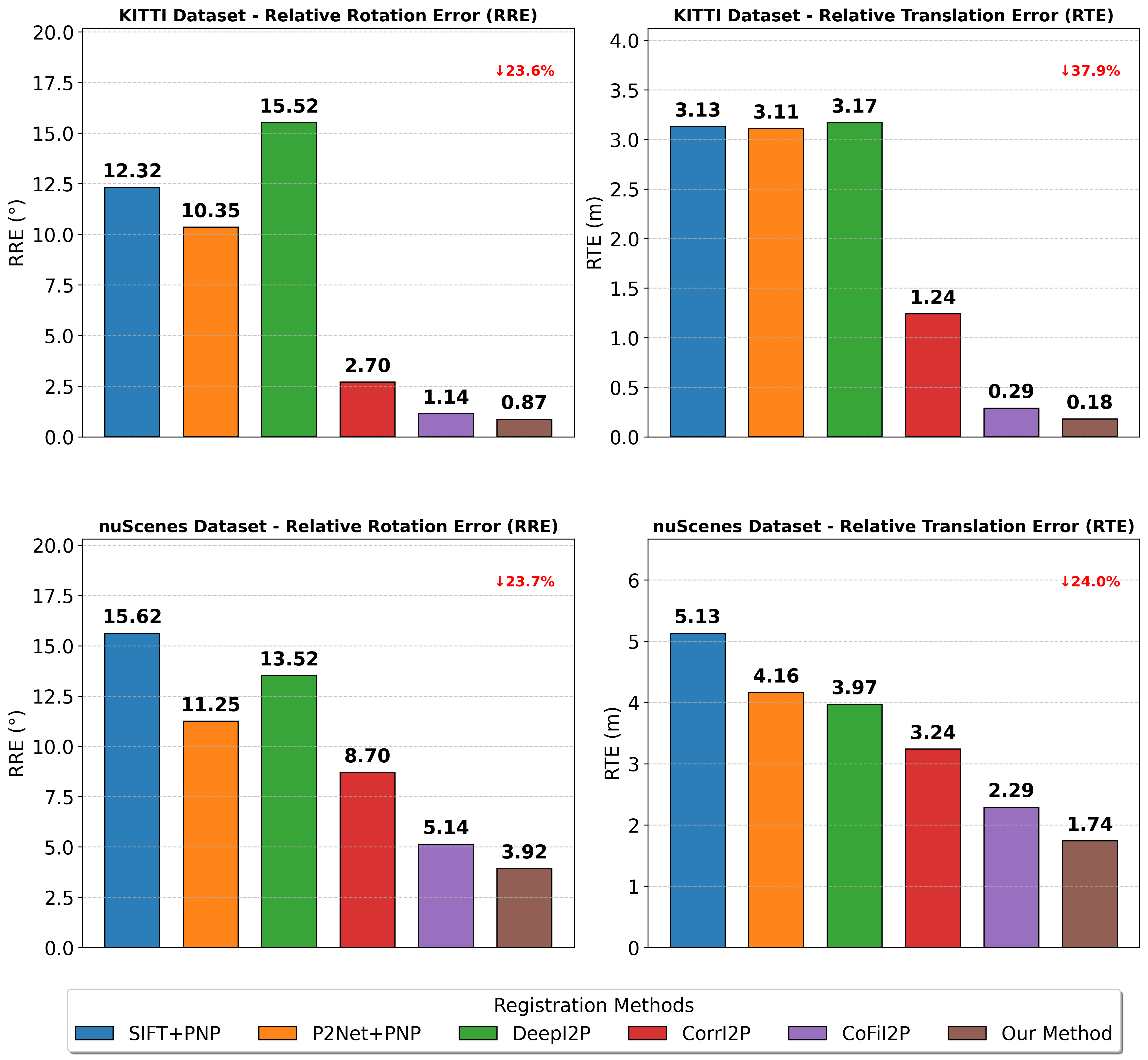}
	\caption{  Performance Comparison of Registration Methods.    }
	\label{fig:error}
\end{figure}
\begin{figure}[!htb]
	\centering 
	\includegraphics[width=1.0\hsize]{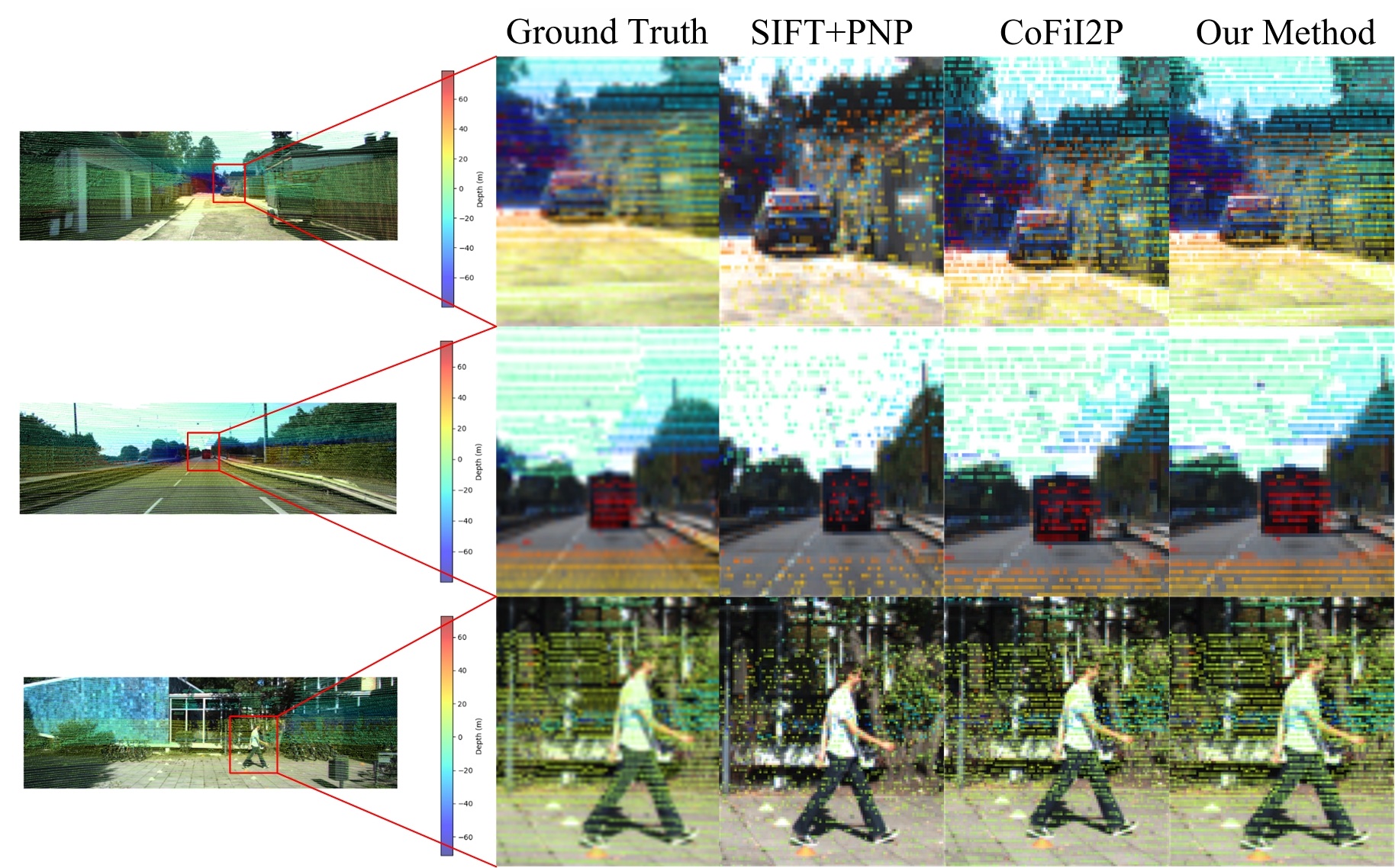}
	\caption{ The registration effect of different methods on the dataset, from which we can observe that our method obtains better alignment results.}
	\label{fig:View}
\end{figure}

Figure 5 shows the visualization results of the algorithm. We compared the ground truth Traditional method (Sift+PNP), CoFiI2P and our method. Different colors indicate different depths of point clouds. It can be clearly seen that our method obtains more matching points and more accurate projections.

In order to understand the loss changes of multiple modules during the training process, we defined the overall loss, self-supervised modal loss, coarse registration loss, and fine registration loss as Total Loss, Description Loss, Coarse Loss, and Fine Loss, and analyzed them on the Nuscence dataset. The results are shown in Figure 6. It can be seen from this that the overall training loss and self-supervised modal loss are continuously decreasing. The coarse registration loss first rapidly decreases, and after stabilization, the fine registration loss begins to decrease, proving the regulatory effect of the DC module.

\begin{figure}[!htb]
	\centering 
	\includegraphics[width=1\hsize]{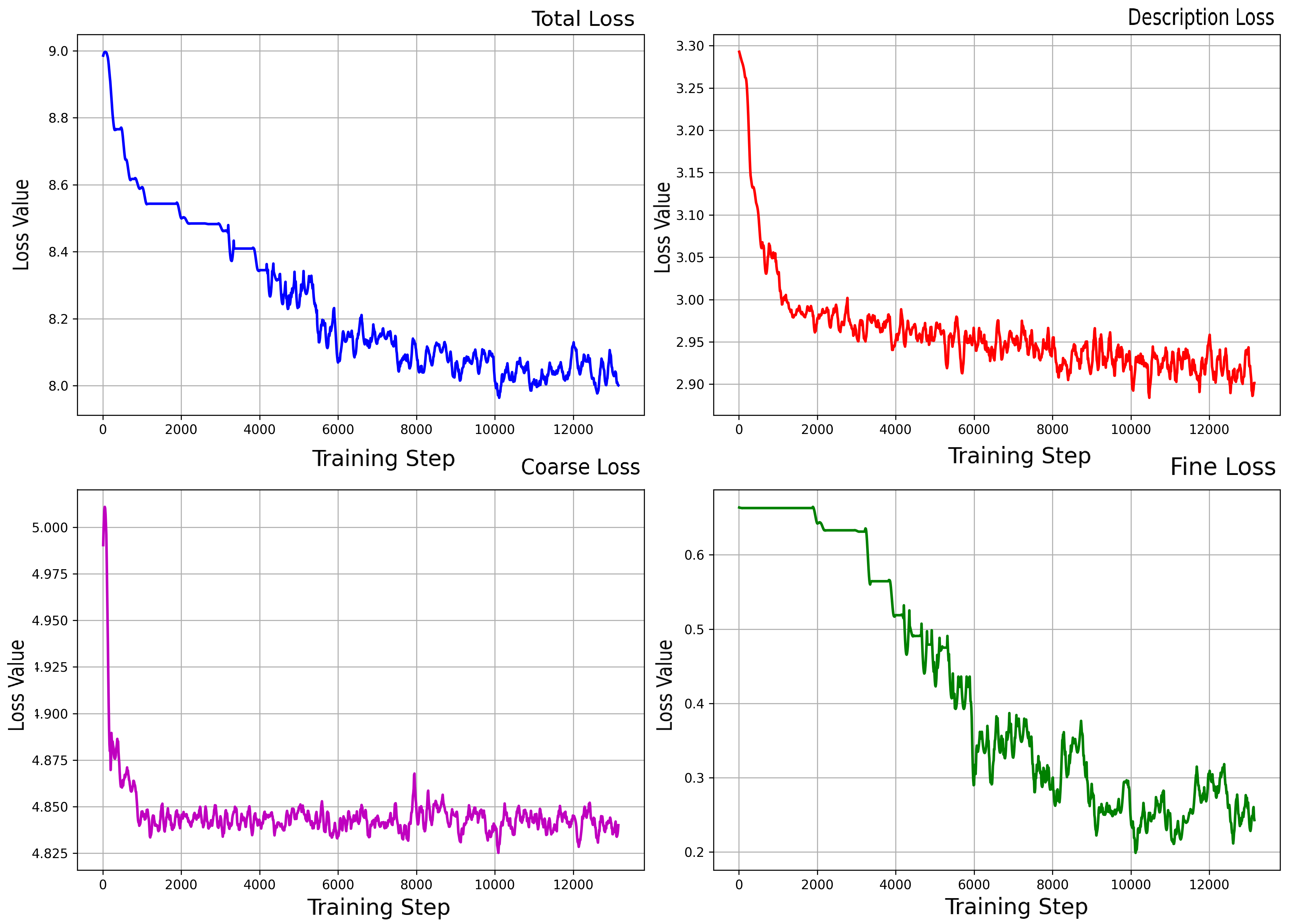}
	\caption{ Changes in losses during the training process.}
	\label{fig:training_losses}
\end{figure}

Finally, to verify the stability of the algorithm under noise and occlusion conditions, we conducted experiments on the nuScence dataset, and the experimental results are shown in Table \ref{tab:dataset_stats}. Among them, noise\_0.1\_occlusion\_16x16 represents a point cloud Gaussian noise standard deviation of 0.1, and the image occlusion area size is 16x16. From the table, it can be seen that our method can still maintain good performance even in situations where noise and occlusion are quite severe.

\begin{table}[ht]
\small
\centering
\caption{Experimental Result on nuScence.}
\label{tab:dataset_stats}
\begin{tabular}{lccc}
\toprule
\textbf{Noise setting} & \textbf{RRE($^\circ$)} & \textbf{RTE(m)} & \textbf{IR(\%)}  \\
\midrule
noise\_0.0\_occlusion\_0x0       & 3.92±2.29           & 1.74±0.98       & 82.58              \\

noise\_0.0\_occlusion\_16x16     & 4.02±2.31         & 1.78±0.95
           & 80.35                      \\

noise\_0.0\_occlusion\_32x32       & 4.06±2.34           & 1.77±0.95            & 79.56                     \\

noise\_0.1\_occlusion\_0x0       & 3.95±2.22           & 1.79±0.96          & 79.34                      \\

noise\_0.1\_occlusion\_16x16       & 4.02±2.38          & 1.82±0.93
        &  77.75            \\

noise\_0.1\_occlusion\_32x32       & 3.98±2.30      &1.84±0.96      &  75.25                     \\
noise\_0.2\_occlusion\_32x32       &  4.21±2.38       &1.94±0.96      &  73.74     \\
\bottomrule
\end{tabular}
\end{table}

\subsection{Ablation Study}
In order to comprehensively evaluate the contribution of each core component in CrossI2P, we conducted systematic ablation experiments on four key modules: self-supervised contrastive learning network (SL), two-stage registration feature matching (RM), differential PNP module (D-PNP), and dynamic collaborative training mechanism (DC). The ablation results summarized in Tables 4 and 5 quantitatively demonstrate the indispensability of each module. Below, we will delve into their individual and collaborative impacts on registration performance.

The SL module bridges the geometric semantic gap between 2D images and 3D point clouds through bidirectional comparison alignment. The RM module refines the corresponding relationship from global superpoint/superpixel alignment to local point-level registration. Ablating RM simplifies the framework into single-stage matching, lacking global context awareness. The D-PNP module achieves end-to-end training through differentiable refactoring. The DC module adaptively balances the gradients between feature learning, registration, and pose estimation tasks. Disabling DC will result in fixed loss weights, leading to optimization conflicts. Attitude estimation loss dominates early training, suppresses feature alignment gradients, and leads to suboptimal embedding. 

In summary, ablation studies have rigorously confirmed that the performance of CrossI2P does not come from isolated components, but from their careful integration. Each module addresses different challenges, and together they form a robust framework for uncalibrated cross-modal registration.

\begin{table}[ht]
\small
\centering
\fontsize{9pt}{10pt}\selectfont
\setlength{\tabcolsep}{2.5pt}
\caption{Ablation Experiment Results.}
\begin{tabular}{ccccccccc}
\toprule
\multirow{2}{*}{\textbf{SL}} & 
\multirow{2}{*}{\textbf{RM}} & 
\multirow{2}{*}{\textbf{D-PNP}} & 
\multirow{2}{*}{\textbf{DC}} & 
\multicolumn{2}{c}{\textbf{KITTI}} & 
\multicolumn{2}{c}{\textbf{nuScenes}} \\
\cmidrule(lr){5-6}\cmidrule(lr){7-8}
& & & & 
\textbf{RRE($^\circ$)} & \textbf{RTE(m)} & 
\textbf{RRE($^\circ$)} & \textbf{RTE(m)} \\
\midrule
       & $\checkmark$ & $\checkmark$ & $\checkmark$ & 2.28 & 0.43 & 5.28 & 2.54 \\
\midrule
$\checkmark$ &        & $\checkmark$ & $\checkmark$ & 1.75 & 0.31 & 5.75 & 2.76 \\
\midrule
$\checkmark$ & $\checkmark$ &       & $\checkmark$ & 1.55 & 0.28 & 5.23 & 2.33 \\
\midrule
$\checkmark$ & $\checkmark$ & $\checkmark$ &       & 1.25 & 0.30 & 4.25 & 2.14 \\
\midrule
$\checkmark$ & $\checkmark$ & $\checkmark$ & $\checkmark$ & \textbf{0.87} & \textbf{0.18} & \textbf{3.92} & \textbf{1.74} \\
\bottomrule
\end{tabular}
\end{table}

\section{Conclusion}\label{sec:conclusion}


In this paper, we propose CrossI2P, a unified framework that combines self-supervised learning with a two-stage feature matching strategy to achieve end-to-end 2D–3D registration. CrossI2P resolves the core conflict between semantic alignment and geometric optimization through two key innovations: a self-supervised feature extraction module that bridges the domain gap between sparse LiDAR point clouds and dense RGB images, and a transformer-based coarse-to-fine alignment architecture that avoids suboptimal solutions typically encountered in iterative matching pipelines. Furthermore, the integration of a differentiable PnP module and a dynamic gradient coordination mechanism enables joint optimization of feature learning, correspondence refinement, and pose estimation within a single training loop. In the future, we plan to enhance CrossI2P for real-time deployment via lightweight model distillation and extend its applicability to broader multi-sensor fusion scenarios.

\bibliography{sample-base}


\makeatletter
\@ifundefined{isChecklistMainFile}{
  \newif\ifreproStandalone
  \reproStandalonetrue
}{
  \newif\ifreproStandalone
  \reproStandalonefalse
}
\makeatother

\ifreproStandalone
\documentclass[letterpaper]{article}
\usepackage[submission]{aaai2026}
\setlength{\pdfpagewidth}{8.5in}
\setlength{\pdfpageheight}{11in}
\usepackage{times}
\usepackage{helvet}
\usepackage{courier}
\usepackage{xcolor}
\frenchspacing

\begin{document}
\fi
\setlength{\leftmargini}{20pt}
\makeatletter\def\@listi{\leftmargin\leftmargini \topsep .5em \parsep .5em \itemsep .5em}
\def\@listii{\leftmargin\leftmarginii \labelwidth\leftmarginii \advance\labelwidth-\labelsep \topsep .4em \parsep .4em \itemsep .4em}
\def\@listiii{\leftmargin\leftmarginiii \labelwidth\leftmarginiii \advance\labelwidth-\labelsep \topsep .4em \parsep .4em \itemsep .4em}\makeatother

\setcounter{secnumdepth}{0}
\renewcommand\thesubsection{\arabic{subsection}}
\renewcommand\labelenumi{\thesubsection.\arabic{enumi}}

\newcounter{checksubsection}
\newcounter{checkitem}[checksubsection]

\newcommand{\checksubsection}[1]{%
  \refstepcounter{checksubsection}%
  \paragraph{\arabic{checksubsection}. #1}%
  \setcounter{checkitem}{0}%
}

\newcommand{\checkitem}{%
  \refstepcounter{checkitem}%
  \item[\arabic{checksubsection}.\arabic{checkitem}.]%
}
\newcommand{\question}[2]{\normalcolor\checkitem #1 #2 \color{blue}}
\newcommand{\ifyespoints}[1]{\makebox[0pt][l]{\hspace{-15pt}\normalcolor #1}}

\ifreproStandalone
\end{document}
\fi

\end{document}